\documentclass[review]{elsarticle}

\usepackage{multicol}%
\usepackage[utf8]{inputenc} 
\usepackage{amssymb}
\usepackage{graphicx}
\usepackage{float}
\restylefloat{table}
\usepackage{placeins}
\usepackage{float}
\usepackage{wrapfig}
\usepackage{algorithm}
\usepackage{algorithmic}
\usepackage{array}
\usepackage{color}
\usepackage{csquotes}
\usepackage{amsmath}
\usepackage{tikz}
\usepackage[utf8]{inputenc}
\usepackage{float}
\usepackage{subfig}
 \usepackage[strict]{changepage}
\usepackage[skip=0pt]{caption}
\usetikzlibrary{positioning,calc}
\usetikzlibrary{shapes,fit} 
\usetikzlibrary{shapes,arrows}
\usepackage{graphicx}
\usepackage{lipsum}
\usepackage{amsfonts,booktabs,multirow,array,comment}
\usepackage{footnote}
\usepackage[hyphens]{url}
\usepackage{gensymb}
\usepackage{color}
\renewcommand*{\arraystretch}{4}
\usepackage{graphbox} 

\usepackage{color}
\usepackage{mathtools}
\usepackage{textcomp}
\definecolor{lightgray}{gray}{0.5}
\makeatletter
\newcommand{\removelatexerror}{\let\@latex@error\@gobble}
\makeatother
\newcolumntype{C}{>{\centering\arraybackslash}p{5em}}

\newcommand{\beq}{\begin{equation}}
\newcommand{\eeq}{\end{equation}}
\newcommand{\bea}{\begin{algorithmic}}
\newcommand{\eea}{\end{algorithmic}}

\usepackage{xparse}
\NewDocumentCommand{\ceil}{s O{} m}{%
  \IfBooleanTF{#1} 
    {\left\lceil#3\right\rceil} 
    {#2\lceil#3#2\rceil} 
}
\usepackage{cleveref}
\crefname{section}{§}{§§}
\Crefname{section}{§}{§§}

\journal{} 

\begin{document}
\begin{frontmatter}
\title{Sparse Filtered SIRT for Electron Tomography}


\author[1]{Chen Mu} 
\author[2]{Chiwoo Park\corref{cor1}}
\cortext[cor1]{Corresponding author: 
  Tel.: +1-850-410-6457;  
  fax: +1-850-410-6342;}
\ead{cpark5@fsu.edu}
\address[1]{PhD Student, Department of Industrial and Manufacturing Engineering, Florida State University, 2525 Pottsdamer St. Tallahassee, FL 32310}
\address[2]{Associate Professor, Department of Industrial and Manufacturing Engineering, Florida State University, 2525 Pottsdamer St. Tallahassee, FL 32310}
\begin{abstract}
Electron tomographic reconstruction is a method for obtaining a three-dimensional image of a specimen with a series of two dimensional microscope images taken from different viewing angles. Filtered backprojection, one of the most popular tomographic reconstruction methods, does not work well under the existence of image noises and missing wedges. This paper presents a new approach to largely mitigate the effect of noises and missing wedges. We propose a novel filtered backprojection that optimizes the filter of the backprojection operator in terms of a reconstruction error. This data-dependent filter adaptively chooses the spectral domains of signals and noises, suppressing the noise frequency bands, so it is very effective in denoising. We also propose the new filtered backprojection embedded within the simultaneous iterative reconstruction iteration for mitigating the effect of missing wedges. Our numerical study is presented to show the performance gain of the proposed approach over the state-of-the-art.
\end{abstract}


\end{frontmatter}
\section{Introduction}
Electron tomographic reconstruction is a method for reconstructing detailed three-dimensional structures of a specimen with a series of two-dimensional transmission electron microscope images of the specimen from different viewing angles, which has been widely used in materials science and biological science \cite{Jinnai2009, Midgley2003}. In practice, a sample stage containing a specimen is tilted around a single axis, and the 2D electron microscope images (\textit{sinograms}) of the specimen are taken for a range of tilt angles with constant intervals. The 3D reconstruction of the specimen (\textit{tomogram}) can be achieved by combining the 2D projection images using a computerized tomography reconstruction algorithm. There are two factors that determine the quality of the reconstruction: (1) the range and interval of tilt angles, and (2) the signal-to-noise ratio (SNR) of sinograms. Due to instrumentation limitation or specimen thickness, the range of the tilt angles is typically limited to $[-65, 65]$ to $[-75, 75]$ degrees, which creates a ``missing wedge'' of information. Reconstruction with the missing wedge typically leaves image artifacts in the reconstruction outcome. In addition, when the signal-to-noise ratio of sinograms is low, the reconstruction outcome could be very noisy. This paper is concerned with improving the accuracy and computational efficiency of electron tomographic reconstruction under those limiting factors.  \\

Past works in tomographic reconstruction are largely categorized into two studies, analytical methods and algebraic methods. Analytical methods are based on a continuous representation of the inverse Radon transform using the Central Slice Theorem \cite{Kak2001} and the computation of the discrete version of the continuous representation. A popular analytical method is the filtered backprojection or shortly FBP \cite{Pan2009}. The FBP is computationally efficient and simple to implement, but it does not work well under a low signal-to-noise (SNR) ratio of input sinograms. The low SNR value is quite common in practice \cite{Hanson1981, Hegerl2006}, and the FBP typically produces a very noisy reconstruction with low SNR sinograms. A quick remedy of this issue is to change the filter function in FBP with a low-pass filter such as Cosine or Hann filter. In addition to the issue, FBP suffers from the missing wedge effects when sinograms are available for only limited tilt angles. \\

Algebraic methods formulate a system of linear equations relating sinograms to the reconstruction outcome and solve the linear equations using an iterative approach, iteratively updating its solution so that it minimizes the mean squared difference between the left and right hand sides of the equations. Depending on their modeling and iterative steps, there are many variants, including the algebraic reconstruction technique \cite[ART]{Gordon1970}, the simultaneous iterative reconstruction technique \cite[SIRT]{Herman1980, Kak2001} and the simultaneous algebraic reconstruction technique \cite[SART]{Andersen1984}. In some literature, the squared difference criterion of the iterative optimization in the algebraic methods was modified to incorporate certain regularization terms such as the total variation regularization \cite{hansen2011total}, the Tikhonov regularization \cite{Vauhkonen1998, Mueller2012} and edge-preserving regularization \cite{Yu2002}, and it has been numerically shown that the regularizations can improve the reconstruction quality. In general, the algebraic methods perform better than the analytical methods for the cases with limited tilt angles, but these advantages come with the expense of higher computational costs. The cost per iteration increases quadratically with the size and number of input sinograms, and the slow convergence of the iterative approach requires many of those expensive iterations. There are some more efficient approaches that implement algebraic methods by using graphic processing units (GPUs) \cite{Flores2013, Xu2005}. In addition, the number of the iterations in the algebraic approaches is a tuning parameter, which typically determines how much detailed features are included in the resulting reconstruction; a too large choice would give a noisy reconstruction, and a too small choice would not give a reconstruction with many missing details. The tuning parameter is generally difficult to determine. \\

Some improvements of the FBP were sought for better dealing with measurement noises and the missing wedge issue, taking the computational advantages of FBP. The simplest approach is the denoising-and-backprojection approach, which applies denoising filters on projection data before the backprojection is performed \cite{Ren2013}. Karimi et. al. \cite{Karimi2016} proposed an improved patch-based denoising algorithm, and a penalized weighted least-squares investigation is made in sinogram denoising for low dose X-ray \cite{Wang2006}. These simple ideas have shown better performance over the FBP for reconstructing with noise sinograms. \\

More recent approaches tried to optimize the filter in FBP with a certain optimization criterion that was originally proposed and used for the algebraic methods. By doing so, the performance improvement of FBP to the accuracy comparable to the algebraic methods was sought. In this paper, we refer those methods as a filter-optimization approach. Zeng \cite{Zeng2012_2} compared the algebraic and FBP approaches, showing that each iteration of the algebraic methods can be seen as `first backprojecting data and filtering the backprojected images with a ramp filter,' and it is equivalent to `first filtering and then backprojecting the filtered data.' The finding became the basis to define a novel FBP that behaves similarly as the algebraic method. Some variants of the approaches \cite{Zeng2012, Zeng2013} were proposed. The major benefits of these approaches are that they are just as fast as the original FBP, while their outcomes are comparable to those of the analytical approaches. The major drawback of the approaches is that the methods come with a tuning parameter $k$ that is related to the degree of denoising, which is typically pre-fixed using the rule of thumb without considering the actual noise-level in data. Pelt proposed a data-pendent FBP that optimizes the FBP filter more adaptively to input data and its noise level \cite{Pelt2014, Pelt2015}, but the approach still requires multiple iterations to optimize the filter. Therefore, the approach does not have much computational gain over the algebraic approaches. \\

We propose a new filter-optimization approach and its embedding within SIRT iterations. We first formulate a new optimization problem of optimizing the filter in FBP with the objective function that comprises a reconstruction accuracy term and a regularization term. The regularization term is the sparsity regularization in a frequency domain to suppress the noisy frequency bands of input sinogram data, and the weighting on the regularization term is determined adaptively to the noise level of the sinogram data. Therefore, the new approach produces a data-dependent filter like Pelt's \cite{Pelt2015}, but the proposed approach does not require any computationally expensive iterations to optimize the filter. Therefore, the proposed approach has computational benefit over the existing data-dependent filters, while maintaining its reconstruction accuracy. We also propose the embedding of the new filter optimization approach with SIRT iterations, replacing the backprojection operator in the SIRT iterations with the new filtered backprojection, which further improves the reconstruction accuracy and reduces missing wedge effects. The integrated algorithm converges significantly faster than the SIRT iterations, because the regularization term in our filter optimization regularizes the sequence of solution updates so draws a faster convergence of the iterations. To sum up, we are proposing two novel approaches: a new filter-optimization approach and its integration with the SIRT. The filter-optimization approach is referred to as the data-dependent sparse filtered backprojection (sFBP), and its integration into the SIRT is referred to as the sparse filtered SIRT (sfSIRT).\\

The remainder of this paper is organized as follows. Section \ref{sec:sfbp} describes the sFBP with its numerical comparison to the state-of-the-art for noisy sinograms. Section \ref{sec:sfsirt} describes the sfSIRT with its numerical comparison to existing iterative construction approaches for limited tilt series. Section \ref{sec:3d} presents the 3D reconstruction examples using experimental tomography data. Finally we have conclusions in Section \ref{sec:con}.

\section{Sparse Filtered Backprojection (sFBP)}\label{sec:sfbp}
In electron tomography, a sample stage is rotated around a tilt axis $z$, and an electron beam irradiates the sample stage in the direction parallel to the $(x,y)$-plane. Therefore, when we regard a 3D tomogram as a stack of its 2D slices along the $(x,y)$-plane, each of the 2D slices can be reconstructed independently using the corresponding slice of sinograms \cite{acar2016adaptive}. In this section, we describe our new filter backprojection approach to reconstruct each 2D slice. Since the $z$ coordinate is fixed for the reconstruction, we are skipping writing the $z$ coordinate in describing our approach for simpler notations. \\


Let $f_0(x,y)$ denote an unknown sample image of interest in (x,y)-space. Let $\mathcal{R} \circ f_0(r, \theta)$ denote the line integral of the image over the line characterized by $x\cos\theta + y\sin\theta = r$, 
\begin{equation} \label{eq:proj}
\begin{split}
\mathcal{R} & \circ f_0(r, \theta) \\ 
= & \int_{-\infty}^{\infty}\int_{-\infty}^{\infty}f_0(x,y)\delta (x\cos{\theta}+y\sin{\theta}-r)\mathrm{d}x\mathrm{d}y.
\end{split}
\end{equation}
When there is no measurement noise, the line integrals for different values of projection dimension $r$ and rotation $\theta$ are measured in tomography, which are referred to as \textit{sinogram}. We denote the non-noisy sinogram by $p_0(r,\theta)$. Tomography reconstruction is to recover $f_0(x,y)$ from the sinogram $p_0(r,\theta)$ or more often its noisy version. \\

 We first introduction a few notations to describe the standard FBP and our sparsity-regularized FBP (sFBP). Let $\mathcal{F}_2 \circ f_0(u,v)$ denote the 2-D Fourier transform of $f_0(x,y)$. Therefore, $f_0(x,y)$ can be recovered by taking the inverse Fourier transform of $\mathcal{F}_2\circ f_0(u,v)$,
\begin{equation*}
f_0(x,y)=\int_{-\infty}^{\infty}\int_{-\infty}^{\infty}\mathcal{F}_2\circ f_0(u,v)e^{j2\pi (ux+vy)}\mathrm{d}u\mathrm{d}v.
\end{equation*}
By the change of variables from $(u, v)$ to $(\omega,\theta)$ with the relation $u=\omega\cos\theta$ and $v=\omega\sin\theta$, we can rewrite the inverse Fourier transform,
\begin{equation} \label{eq:ifr}
\begin{split}
 & f_0(x,y) = \int_{0}^{\pi}\int_{-\infty}^{\infty} q(\omega, \theta) |\omega| e^{j2\pi w(x\cos\theta+y\sin\theta)}\mathrm{d}\omega\mathrm{d}\theta,
\end{split}
\end{equation}
where $q(\omega, \theta) = \mathcal{F}_2 \circ f_0(\omega\cos\theta,\omega\sin\theta)$. The central slice theorem \cite{Kak2001} basically reads
\begin{equation*}
q(\omega, \theta)  = \mathcal{F}_1\circ p_0(\omega, \theta),
\end{equation*}
where $\mathcal{F}_1\circ p_0(\omega, \theta)$ is the 1-D Fourier transform of the sinogram $p_0(r, \theta)$ with respect to the first input dimension $r$. Using the central slice theorem, we re-state equation \eqref{eq:ifr} with the 1-D Fourier transform term as
\begin{equation*} 
 f_0(x,y) =\int_{-\infty}^{\infty}  \int_{0}^{\pi} \mathcal{F}_1\circ {p_0}(\omega, \theta) |\omega|e^{j2\pi w(x\cos\theta+y\sin\theta)}\mathrm{d}\theta \mathrm{d}\omega.
\end{equation*}
The equation basically relates the non-noisy sinogram $p_0(r, \theta)$ to a real space image $f_0(x,y)$ as
\begin{equation*}
f_0(x,y) = \int_{-\infty}^{\infty}\int_{0}^{\pi} Q_{p_0}(\omega;x,y,\theta)|\omega| \mathrm{d}\theta \mathrm{d}\omega.
\end{equation*}
where $Q_{p_0}(\omega;x,y,\theta) = \mathcal{F}_1\circ {p_0}(\omega, \theta) e^{j2\pi w(x\cos\theta+y\sin\theta)}$. When the sinogram data are only available at a finite number of locations $(\omega, \theta)$, so the integration equation is approximated by its finite dimensional version. Let $\Omega$ denote a finite number of the $\omega$ values and $\Theta$ denote a finite set of the $\theta$ values. The finite dimensional version is
\begin{equation}
f_0(x,y) \approx \sum_{\omega \in \Omega} \left\{\sum_{\theta \in \Theta} Q_{p_0}(\omega;x,y,\theta) \right\}|\omega|.
\end{equation}
This defines the linear problem for the filtered backprojection procedure \cite{Kak2001}, which is basically the weighted sum of $Q_{p_0}(\omega;x,y,\theta)$ with weight $|\omega|$; the weight is referred as to the filter of the backprojection, and the specific choice $|\omega|$ is known as the Ram-Lak filter. \\

In practice, the sinogram data $p_0(r,\theta)$ is corrupted through an additive noise process $\epsilon(r, \theta)$,
\begin{equation}
p(r,\theta)= p_0(r,\theta) + \epsilon(r, \theta).
\end{equation}
When the filtered backprojection procedure is applied to the noisy version $p(r,\theta)$, the reconstruction outcome becomes
\begin{equation}
\begin{split}
f(x,y) =&  \sum_{\omega \in \Omega} \sum_{\theta \in \Theta}  Q_{p}(\omega;x,y,\theta)|\omega| \\
       =&  \sum_{\omega \in \Omega} \sum_{\theta \in \Theta}  Q_{p_0}(\omega;x,y,\theta)|\omega| \\
       &  + \sum_{\omega \in \Omega} \sum_{\theta \in \Theta}  Q_{\epsilon}(\omega;x,y,\theta)|\omega| \\
       =& f_0(x, y) + \sum_{\omega \in \Omega}\sum_{\theta \in \Theta}  Q_{\epsilon}(\omega;x,y,\theta)|\omega|,
\end{split}
\end{equation}
where $Q_{\epsilon}(\omega;x,y,\theta) = \mathcal{F}_1\circ {\epsilon}(\omega, \theta) e^{j2\pi w(x\cos\theta+y\sin\theta)}$ is the effect of observation noise on the reconstructed image $f(x,y)$. Note that the noise effect term, $Q_{\epsilon}(\omega;x,y,\theta)$, is summed over different frequencies $\omega$. We assume that there are some frequency bands $\omega$ with relatively low $Q_{\epsilon}(\omega;x,y,\theta)$ and relatively high $Q_{p_0}(\omega;x,y,\theta)$; otherwise, noises and true signals are indistinguishable in the Fourier domain. Let $W$ denote the set of the frequency bands. We propose to find $W$ and suppress $Q_{\epsilon}(\omega;x,y,\theta)$ for $\omega \notin W$ by replacing the filter $|\omega|$ with
\begin{equation} \label{eq:filterform}
|\omega| \delta(\omega \in W),  W \subset \Omega,
\end{equation}
where $\delta(\cdot)$ is the Dirac delta function. The modification is valued $|\omega|$ when $\omega$ is an element of the subset $W$ and is zero-valued otherwise. The choice of the subset $W$ is important. Ideally, $W$ should exclude the frequency bands dominated by noises. Let $f_W(x,y)$ denote the resulting reconstruction for a choice of $W$, 
\begin{equation} \label{eq:fourier_W}
\begin{split}
f_W(x, y) = \sum_{\theta \in \Theta} \sum_{\omega \in \Omega} & (\mathcal{F}_1\circ {p}(\omega, \theta)\delta(\omega \in W)) \\
          & \quad e^{j2\pi w(x\cos\theta+y\sin\theta)} |\omega| .
\end{split}
\end{equation}
We optimize the choice of the subset $W$ by minimizing the regularized reconstruction error 
\begin{equation} \label{eq:optm}
\begin{split}
& \mbox{Minimize} \quad ||f_W -  f||_2^2 + \lambda\mathcal{P}(f_W),\\
& \mbox{subject to} \quad W \subset \Omega,
\end{split}
\end{equation}
where $||f||_2^2 := \int\int f^2 dxdy$ is the L2 norm of function $f$, and $\mathcal{P}(f_W) := |W|$ is the cardinality of $W$. The optimization criterion basically pursues for minimizing the reconstruction error while using the information from a less number of the frequency bands. When the value of $f$ is only available at a finite number of $(x,y)$'s, the norm is the sum of the squares of the function values at the finite $(x,y)$ locations. Since the discrete Fourier transform is an orthonormal transformation, the L2 norm of $f$ is equivalent to the sum of the squares of the discrete Fourier coefficients of the function:
\begin{equation}
\begin{split}
||f||_2^2 & = \sum_{\omega \in \Omega, \theta \in \Theta} |\mathcal{F}_2 \circ f(\omega\cos\theta,\omega\sin\theta)|^2.
\end{split}
\end{equation}
Since $\mathcal{F}_2 \circ f(\omega\cos\theta,\omega\sin\theta) = \mathcal{F}_1\circ p(\omega, \theta)$ by the central slice theorem \cite{Kak2001}, the norm is equivalent to
\begin{equation}
\begin{split}
||f||_2^2 & = \sum_{\omega \in \Omega, \theta \in \Theta} |\mathcal{F}_1\circ p(\omega, \theta)|^2.
\end{split}
\end{equation}
Since 
\begin{equation*}
\begin{split}
f_W(x, y) &  - f(x, y) = \\
       & \sum_{\theta \in \Theta} \sum_{\omega \in \Omega} (\mathcal{F}_1\circ {p}(\omega, \theta)\delta(\omega \in W) - \mathcal{F}_1\circ {p}(\omega, \theta)) \\
          & \quad e^{j2\pi w(x\cos\theta+y\sin\theta)} |\omega|,
\end{split}
\end{equation*}
we have
\begin{equation}
\begin{split}
||f_W - f||_2^2 & = \sum_{\omega \in \Omega, \theta \in \Theta} |\mathcal{F}_1\circ p(\omega, \theta) \delta(\omega \in W) \\
              & \qquad \qquad \qquad - \mathcal{F}_1\circ p(\omega, \theta)|^2 \\
              & =  \sum_{\omega \in \Omega\backslash W} \sum_{\theta \in \Theta} |\mathcal{F}_1\circ p(\omega, \theta)|^2,
\end{split}
\end{equation}
and the optimization problem in equation \eqref{eq:optm} becomes
\begin{equation} \label{eq:optm_sim}
\begin{split}
\mbox{Minimize} \quad  & \sum_{\omega \in \Omega\backslash W} \sum_{\theta \in \Theta} |\mathcal{F}_1\circ p(\omega, \theta)|^2 + \lambda|W| \\
\mbox{subject to} \quad & W \subset \Omega.
\end{split}
\end{equation}
One can easily show that the optimal solution for the problem is simply the following hard thresholding rule,
\begin{equation} \label{eq:filter}
W^* = \left\{ \omega \in \Omega; \sum_{\theta \in \Theta} |\mathcal{F}_1\circ p(\omega, \theta)|^2 \ge \lambda   \right\}.
\end{equation}
Applying the optimal choice $W^*$ to the backprojection \eqref{eq:fourier_W} is led to the new approach, data-dependent sparse filtered backprojection (sFBP), which is described by
\begin{equation} \label{eq:fourier_W*}
\begin{split}
f_{W^*}(x, y) = \sum_{\theta \in \Theta} \sum_{\omega \in \Omega} & (\mathcal{F}_1\circ {p}(\omega, \theta)\delta(\omega \in W^*)) \\
          & \quad e^{j2\pi w(x\cos\theta+y\sin\theta)} |\omega| .
\end{split}
\end{equation}

\subsection{Tuning parameter $\lambda$}
The choice of $\lambda$ in the optimization problem \eqref{eq:optm} is critical for the performance. Too large $\lambda$ choices would zero out too many frequency bands so would lead to too much loss in reconstruction details, while too small $\lambda$ choices would keep significant noise information in tomography reconstruction. Mainly motivated by the fact that the optimization formulation \eqref{eq:optm} is a $L1$-penalized regression problem, we considered and tried the existing penalty parameter selection criteria for a L1-penalized regression problem, including Akaike Information Criterion (AIC) \cite{Burnham2002, Akaike1974}, Bayesian Information Criterion (BIC) \cite{Schwarz1978}, and the generalized model description length (gMDL) model selection criterion \cite{Hansen2001, Buhlmann2006}. Among the three, gMDL numerically worked best for most of our numerical cases, so we propose to choose $\lambda$ using gMDL, which chooses the $\lambda$ that minimizes the following generalization error:
\setlength{\arraycolsep}{0.0em}
\begin{eqnarray} \label{eq:gMDL}
gMDL(\lambda)&{}={}&\frac{|\Omega|}{2}\log(||f_W - f||_2)\nonumber\\
&&{+}\:\frac{|W|}{2}\log{\frac{||f_W||_2/|W|}{||f_W - f||_2/{(|\Omega|-|W|)}}}\nonumber\\
&&{+}\log{|\Omega|}.
\end{eqnarray}
The gMDL uses a data driven penalty to the L2-loss in a penalized regression problem, and it overcomes the limitations of the AIC and BIC criteria \cite{Hansen2001, Buhlmann2006}.

\subsection{Implementation and Complexity} 
Suppose that we have noisy sinogram, $p(r, \theta)$, for $r \in \{r_1,r_2,\ldots, r_{N_d}\}$ and $\theta \in \{\theta_1, \theta_2,\ldots, \theta_{N_{\theta}}\}$. Its 1-D fourier transform with respect to dimension $r$ can be taken for each $\theta$ to get 
\begin{equation*}
\mathcal{F}_1\circ p(\omega, \theta).
\end{equation*}
According to \eqref{eq:filter}, the solution for the optimization \eqref{eq:optm_sim} can be achieved by the simple thresholding rule on $\sum_{\theta \in \Theta} |\mathcal{F}_1\circ p(\omega, \theta)|^2$ with threshold $\lambda$, and the possible choices of $\lambda$ are limited to the possible values of $\sum_{\theta \in \Theta} |\mathcal{F}_1\circ p(\omega, \theta)|^2$. For $\lambda$ in the possibles, one can evaluate $gMDL(\lambda)$, and the optimal $\lambda$ is chosen as one achieving the minimum gMDL. Once $\lambda$ is chosen, we can simply evaluate the backprojection \eqref{eq:fourier_W*}. The computation steps of the sFBP is summarized below:
\begin{description}
 \item[\textbf{Input.}]  \quad Discrete projection data $p(r, \theta)$ for $r \in \{r_1,r_2,\ldots, r_{N_d}\}$ and $\theta \in \{\theta_1, \theta_2,\ldots, \theta_{N_{\theta}}\}$. 
 \item[\textbf{Step 1.}]  \quad For each $\theta_j$, take the 1-D Fast Fourier transform of $p(r, \theta=\theta_j)$ and denote the Fourier coefficients by $c(\omega_i, \theta_j) := \mathcal{F}_1\circ {p}(\omega_i, \theta_j)$ for $i=1,...,N_d$ and $j=1,...,N_{\theta}$. 
 \item[\textbf{Step 2.}]  \quad Compute $\alpha_i = \sum_{j=1}^{N_{\theta}} c(\omega_i, \theta_j)^2$. Compute the order statistics of the $\alpha_i$'s by sorting them. Let $i(k)$ denote the index of the $k$th largest one among them. 
 \item[\textbf{Step 3.}]  \quad Denote $k* = \arg\min_{k =1,...,N_d}  gMDL(\lambda = \alpha_{i(k)})$.
 \item[\textbf{Step 4.}]  \quad $\lambda = \alpha_{i(k*)}$ and $W^* = \{ \omega_{i(1)}, \omega_{i(2)}, \ldots, \omega_{i(k*)} \}$.
 \item[\textbf{Step 5.}]  \quad Compute $f_{W^*}$ using equation \eqref{eq:fourier_W*}. 
\end{description}
The computational complexity of Step 1 through Step 4 is dominated by the complexity of Step 1, which is $O(N_{\theta}N_d\log(N_d))$, because the complexity of the 1D fourier transform for each $\theta$ is $O(N_d\log(N_d))$. The complexity of Step 5 is $O(N_{\theta}N^2)$, where we assumed that the backprojection outcome consists of $N \times N$ pixels. When $N_d \approx N$, the total computation complexity is $O(N_{\theta}N^2)$, which is same as the complexity of the standard FBP. 

\subsection{Numerical Comparison of sFBP to the State-of-the-Art} \label{sec:numerical1}
In this section, we present the numerical performance of sFBP for simulated datasets and compare it with those from some baseline methods and the state-of-the-art. For the baseline methods, we chose the FBP with Ram-Lak Filter or shortly FBP(RL) and the FBP with the Hann filter or shortly FBP(HN) to see the performance gain of our filter optimization approach over the standard FBP approaches. Using the Ram-Lak filter typically magnifies high frequency noises in projection data during its reconstruction procedure, while the Hann filter has some ability in filtering out high-frequency noises. We also included the simultaneous iterative reconstruction technique (SIRT) as another baseline method. The performance of the SIRT depends on the number of iterations. We set the numbers of iterations to either 50 or 100. We use SIRT(k) to represent SIRT with $k$ iterations. For the state-of-the-art methods, we chose one representative from each different approach. The iterative and Landweber filtered backprojection \cite[L-FBP]{Zeng2012} was chosen as a representative of the filter optimization approach, which outperformed other filter optimization approaches in our numerical cases. The generalized Tikhonov regularization reconstruction method \cite[TR]{Mueller2012} was chosen as a representative of an algebraic approach. The sinogram denoising method \cite[SMF-FBP]{Qu2012} was chosen as a representative of the sinogram denoising approach, which outperformed other similar approaches. We implemented all the algorithms in Matlab (MathWorks, Inc., MA, USA). \\

\begin{figure}[h]
\centering
\includegraphics[width=\linewidth]{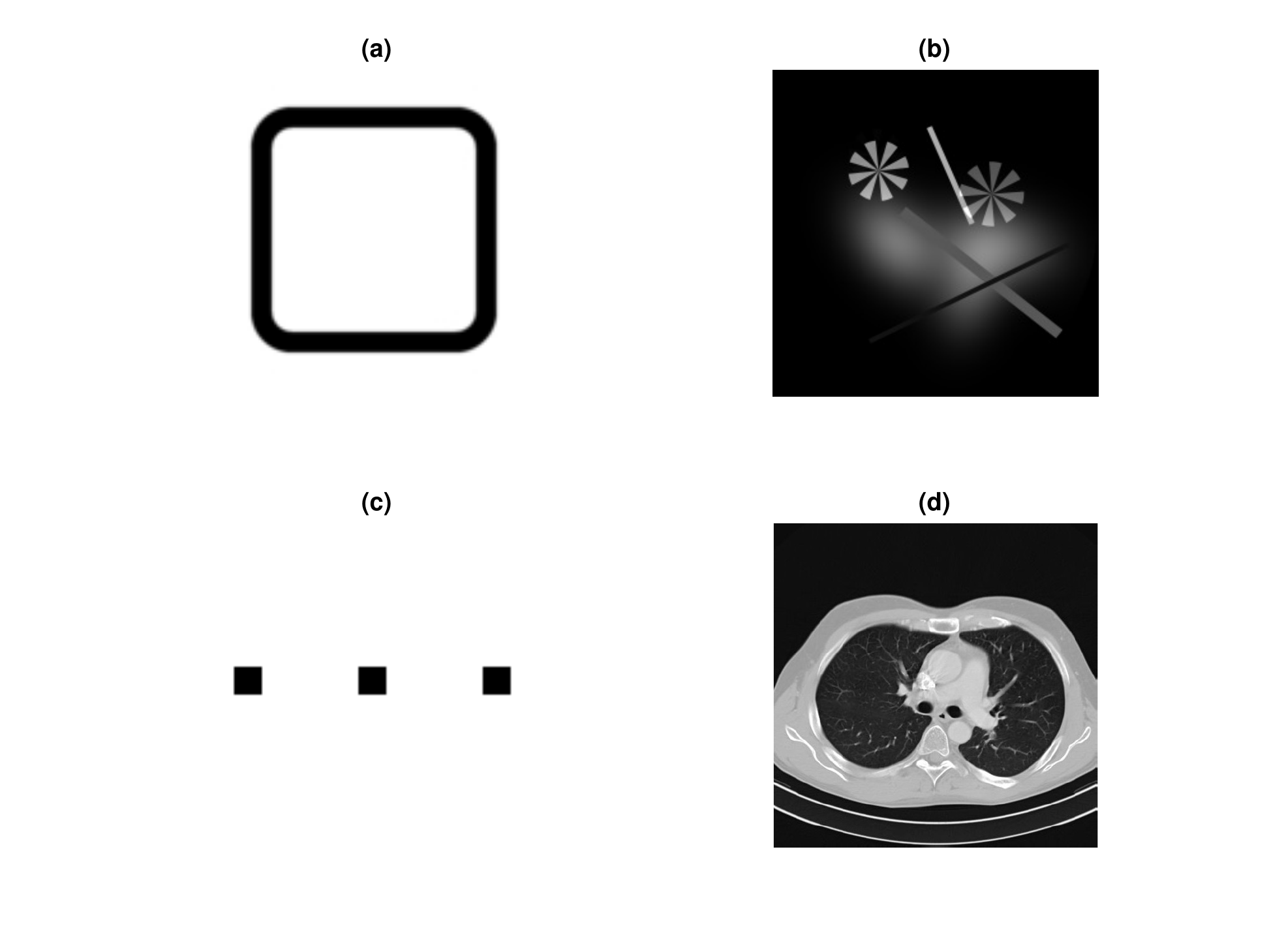}
\caption{Four test images of (a) box phantom image of size $128\times 128$ (b) phantom image of size $256 \times 256$, (c) three-dot phantom image of size $128\times 128$ (d) thorax CT image of size $512\times 512$. They are all of gray scale.} \label{fig:testimages}
\end{figure}

\begin{figure}[h]
\begin{center}
\includegraphics[width=\textwidth]{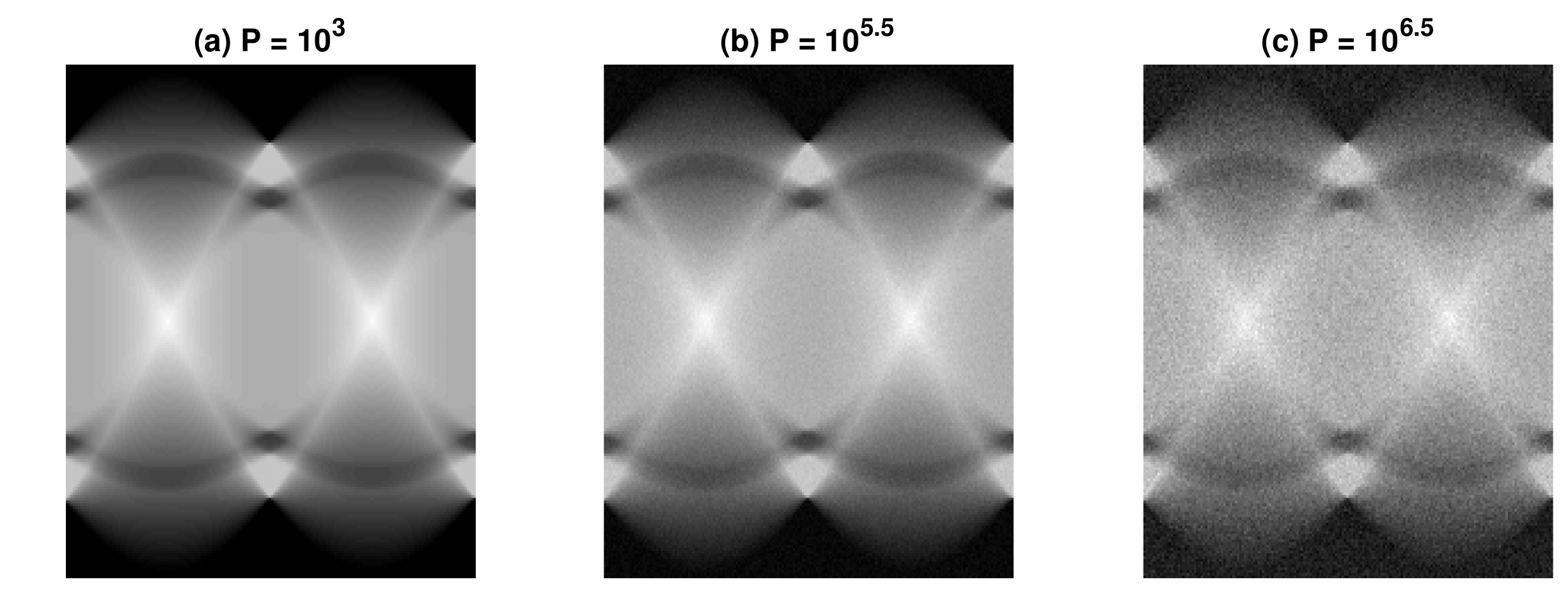}
\caption{Sinogram with Poisson noises; $P$ is the Poisson intensity.} \label{fig:sinogram}
\end{center}
\end{figure}

We compare the different approaches in terms of reconstruction accuracy and computation speed under different simulated scenarios. For the comparison, we used four test images as shown in Figure \ref{fig:testimages}, some of which were from the literature \cite{Pelt2014, thorax}. The sinograms of the four test images were computed using the Radon transform, and the projection data were modified by adding observation noises, before being used in the reconstruction methods. We used Poisson noises, and we varied the Poisson intensity (denoted by $P$) over $10^3$, $10^{5.5}$ and $10^{6.5}$ to simulate different levels of noises. See Figure \ref{fig:sinogram} for exemplary projection data with different noise intensities applied on the second test image. The noisy sinogram data were used as inputs to each of the compared methods, and the resulting reconstruction $\hat{f}$ was compared with the groundtruth $f_0$ to evaluate the reconstruction accuracy. We used two accuracy measures popularly used in computational tomography, the peak signal-to-noise ratio (PSNR) and the structural similarity index (SSIM). The PSNR criterion measures the average reconstruction error:
\begin{equation*}
PSNR(\hat{f}_0)=10\log_{10} (f_{0_{MAX}}^2 / MSE(\hat{f}_0)),
\end{equation*} 
where
\begin{equation*}MSE(\hat{f}_0)=\frac{1}{|X||Y|}\sum_{x \in X}\sum_{y \in Y} (\hat{f}_0(x,y) - f_0(x,y))^2
\end{equation*} 
and $f_{0_{MAX}}$ is the max value among all entries of $f_0(x,y)$.
The SSIM \cite{Wang2004} is a common criterion to quantify the similarity of reconstruction $\hat{f}_0$ and ground truth $f_0$ images in terms of image structures and edge-preserving properties, which is defined by 
\begin{equation*}
SSIM(\hat{f}_0)=[a(f_0,\hat{f}_0)]^\alpha\cdot[b(f_0,\hat{f}_0)]^\beta\cdot[c(f_0,\hat{f}_0)]^\gamma,
\end{equation*} 
where $a(f_0,\hat{f_0})=\frac{2\mu_{f_0}\mu_{\hat{f_0}}+C_1}{\mu_{f_0}^2+\mu_{\hat{f_0}}^2+C_1}$, $b(f_0,\hat{f}_0)=\frac{2\sigma_{f_0}\sigma_{\hat{f_0}}+C_2}{\sigma_{f_0}^2+\sigma_{\hat{f_0}}^2+C_2}$ and $c(f_0,\hat{f}_0)=\frac{\sigma_{f_0\hat{f_0}}+C_3}{\sigma_{f_0}\sigma_{\hat{f_0}}+C_3}$, $\mu_{f_0}$ and $\sigma_{f_0}$ are the local mean and standard deviation for $f_0$, and $\sigma_{f_0\hat{f_0}}$ is the cross-covariance of $\hat{f_0}$ and $f_0$. As suggested in \cite{Wang2004} , we set $\alpha=1$, $\beta=1$ and $\gamma=1$, $C_1 = (0.01L)^2$, $C_2 = (0.03L)^2$, and $C_3 = C_2/2$, where $L$ is a specified dynamic range value of the input image. \\

\begin{figure}[p]
\begin{center}
\includegraphics[width=1.2\textwidth]{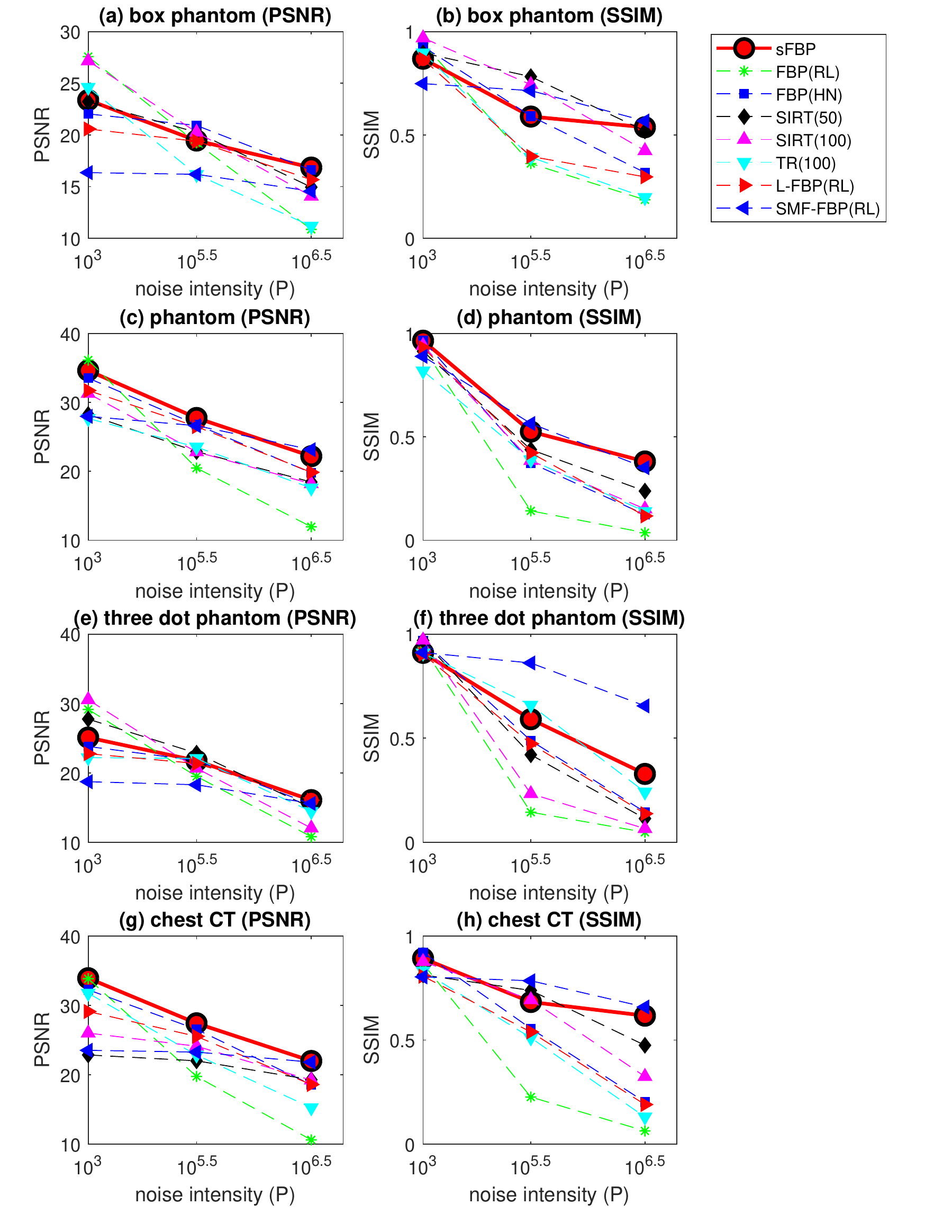} 
\caption{PSNR and SSIM of reconstructions for noisy simulated cases.}\label{fig:test1_psnr}
\end{center}
\end{figure}

\begin{figure}
\begin{center}
\includegraphics[height=0.45\textheight]{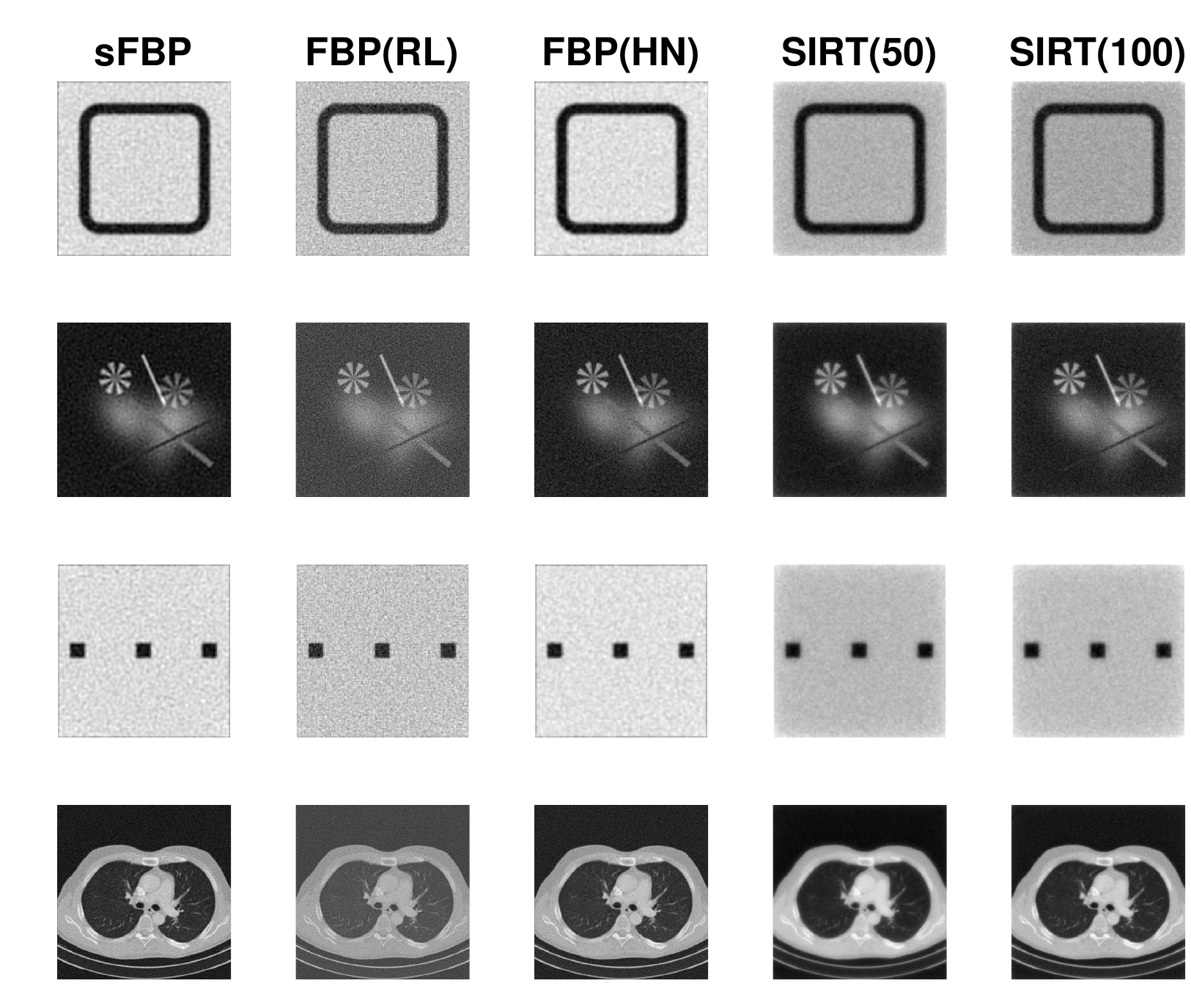} \includegraphics[height=0.45\textheight]{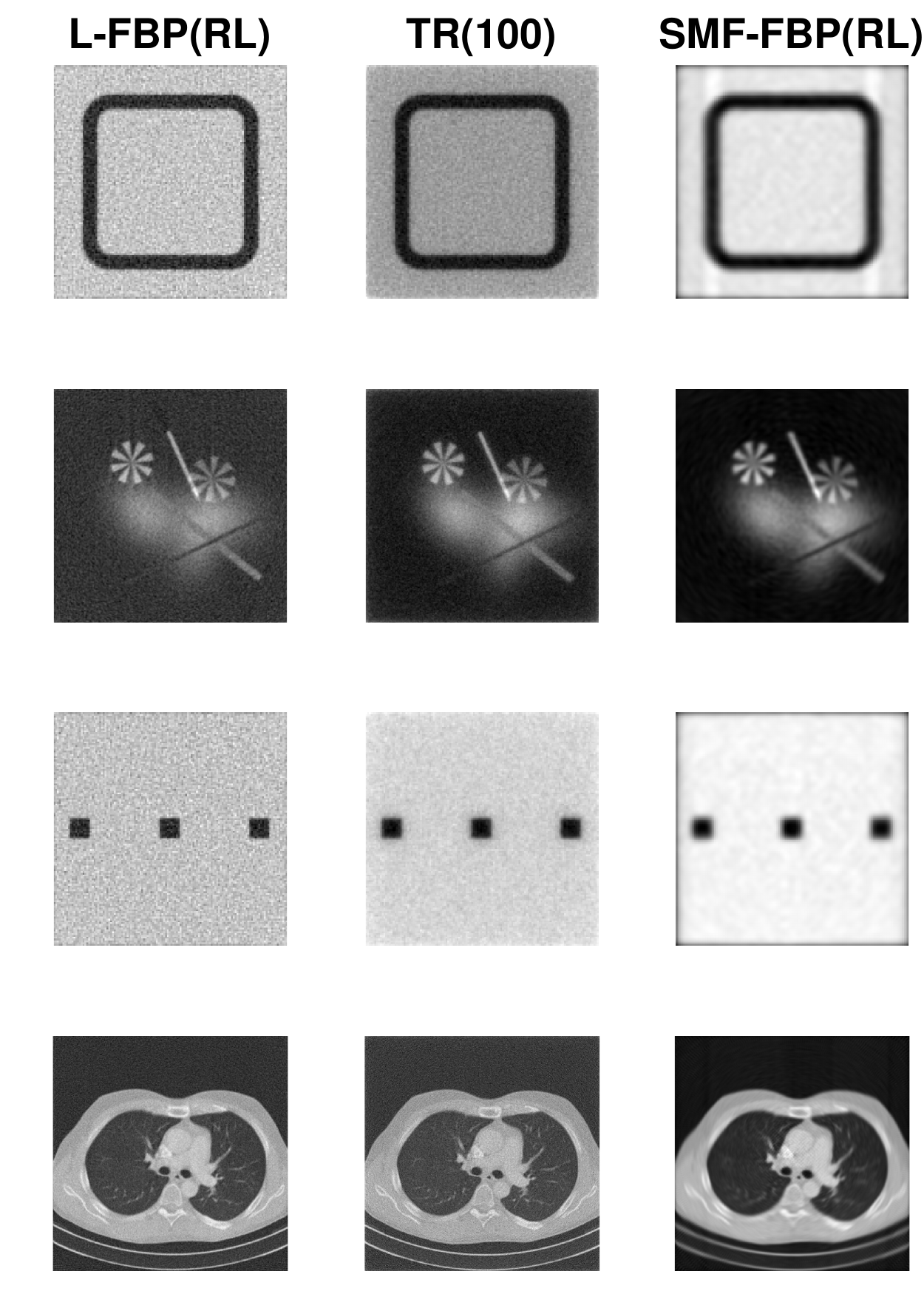}
\caption{Results with all the four test images. We show the reconstruction results from the sFBP, FBP(RL), FBP(HN), SIRT(50), SIRT(100), L-FBP(RL), TR(100) and SMF-FBP methods with P level $10^{5.5}$.}\label{fig:test1}
\end{center}
\end{figure}

Figure \ref{fig:test1_psnr} shows the PSNR and SSIM values for the noisy simulated cases under different noise intensity levels. The PSNR and SSIM values are averaged over ten replicated experiments to reduce the random effect of noise generation. To ease the comparison, we also computed the PSNR and SSIM rank scores of the compared methods. The rank score of a method for each performance metric was computed as a sum of all the ranks among the compared methods over four test images, and lower average rank values implies better ranked. Table \ref{tab:rank1} show the PSNR rank scores and SSIM rank scores. The rank scores vary depending on the noise intensity $P$, but the overall performance of the sFBP is better than the other compared methods. Details of the individual comparison and discussion, based on Figure \ref{fig:test1}, are described below.\\

\begin{table}[h]
\renewcommand{\arraystretch}{1.3}
\subfloat[PSNR]{\begin{tabular}{c|clclclclc}
    \hline
     Noise Intensity & $10^3$ & $10^{5.5}$ & $10^{6.5}$\\
    \hline
   \hline
\begin{minipage}[t]{0.7in}sFBP\end{minipage} & 11	&10&	5\\
\hline
\begin{minipage}[t]{0.7in}FBP(RL)\end{minipage} & 6	& 29	& 32\\
\hline
\begin{minipage}[t]{0.7in}FBP(HN)\end{minipage} &17	&8	&16\\
\hline
\begin{minipage}[t]{0.7in}SIRT(50)\end{minipage} &22	&17&	17\\
\hline
\begin{minipage}[t]{0.7in}SIRT(100)\end{minipage} & 14	&19	& 23\\
\hline
\begin{minipage}[t]{0.7in}TR(100)\end{minipage} & 22	&21&	17\\
\hline
\begin{minipage}[t]{0.7in}L-FBP\end{minipage} & 22&	17	&13\\
\hline
\begin{minipage}[t]{0.7in}SMF-FBP\end{minipage} & 30&	23	&11\\
\hline
\end{tabular}}
\quad
\subfloat[SSIM]{\begin{tabular}{c|clclclclc}
    \hline
     Noise Intensity & $10^3$ & $10^{5.5}$ & $10^{6.5}$\\
    \hline
   \hline
\begin{minipage}[t]{0.7in}sFBP\end{minipage} & 16	&14&	7\\
\hline
\begin{minipage}[t]{0.7in}FBP(RL)\end{minipage} & 15	& 32	& 32\\
\hline
\begin{minipage}[t]{0.7in}FBP(HN)\end{minipage} &8	&20	&20\\
\hline
\begin{minipage}[t]{0.7in}SIRT(50)\end{minipage} &19	&12&	15\\
\hline
\begin{minipage}[t]{0.7in}SIRT(100)\end{minipage} & 8	&18	& 19\\
\hline
\begin{minipage}[t]{0.7in}TR(100)\end{minipage} & 24	&21&	22\\
\hline
\begin{minipage}[t]{0.7in}L-FBP\end{minipage} & 26&	21	&24\\
\hline
\begin{minipage}[t]{0.7in}SMF-FBP\end{minipage} & 28&	6	&5\\
\hline
\end{tabular}}
\caption{Rank scores of the compared methods for different noise intensity levels. Panel (a) contains the rank scores with respect to PSNR, and panel (b) contains the rank scores with respect to SSIM.} \label{tab:rank1}
\end{table}

\textbf{Comparison to FBP.} The FBP(RL) worked very well for the lowest noise intensity level, but it did not work well for higher noise cases, and changing the filter to Hann filter, resulting in FBP(HN), made significant improvement in both PSNR and SSIM. The proposed method, sFBP, worked better than FBP(HN), which justifies the need of the filter optimization proposed in sFBP. The computation times of the two FBP methods and sFBP were comparable, faster than the other competing methods. \\

\textbf{Comparison to SIRT and TR.} Both of the SIRT and TR are algebraic iterative methods, where the TR basically introduces the generalized Tikhonov regularization to modify the SIRT iterations. The SIRT(100) worked better than the SIRT(50) for the lowest noise case with $P=10^{3}$, while the result was opposite for the highest noise case with $P=10^{6.5}$. Please note that the number of iterations typically determines how much detailed features are included in the resulting reconstruction. For the lowest noise case, more iterations would improve the reconstruction accuracy. However, for the highest noise case, more iterations also imply the overfit of the reconstruction to noises, which is also observed in other literature \cite{Zeng2012_2}. The TR method worked comparably to SIRT(50). \\

\textbf{Comparison to L-FBP.} The L-FBP is a filter optimization approach that improves the performance of FBP comparably to the algebraic iterative algorithm and runs as fast as the FBP. It has a tuning parameter, $k$, which controls noise in the reconstruction. In general, large $k$ gives more noisy reconstruction but keeps fine details of image features, while smaller $k$ has  opposite effects. We chose $k = 10000$, which gave best PSNR among the values ranging in 20 to 10000. For the high noise case, the L-FBP performed better than both of FBP(RL) and FBP(HN), but the proposed sFBP method worked better than the L-FBP. \\

\textbf{Comparison to SMF-FBP.} The SMF-FBP \cite{Qu2012} is a sinogram denoising technique that mainly applies a special mean filter on sinogram for denoising and then perform the conventional filtered backprojection on the denoised data. The SMF-FBP performed very well for the high noise case as competitive as the proposed sFBP. However, its performance was not very good for the low and medium noise cases, because it produces undesirable artifact. For the first test image, the SMF-FBP produces two white vertical lines that do not exist in the original image and therefore the PSNR and SSIM are relatively low. For the second image, it produces circular line patterns.

\section{SIRT + sFBP: Sparse Filtered SIRT (sfSIRT)}\label{sec:sfsirt}
In most electron tomography experiments, it is often difficult to obtain 2D sinograms for a full range of 180 degree tilt angles, and the accessible tilt angles typically range from -65 (or 75) degrees to +65 (or +75) degrees for electron tomography experiments. Since many tomography construction approaches were developed assuming that sinogram data are available for a full angular range (i.e. $[-90, 90]$ degrees), applying the approaches with limited angles could create some image artifacts in the resulting reconstruction, which is known as the missing wedge effect. According to our numerical experiments, most of the advanced FBP methods including our sFBP suffers from the missing wedge effect up to different degrees.  To mitigate the missing wedge effect in sFBP, we propose to combine sFBP with the simultaneous iterative reconstruction method. This section describes the idea of integrating sFBP and SIRT and presents its numerical performance. \\

The SIRT algorithm updates its tomographic reconstruction iteratively until the backprojection of its latest reconstruction reaches close to the input projection data. Let $f^{(k)}$ denote the reconstruction achieved at the $k$th iteration of SIRT and let $p$ represents the input sinogram. The $(k+1)$th iteration involves the update,   
\begin{equation*}
f^{(k+1)}=f^{(k)}+\lambda \mathcal{A}^T(p- \mathcal{A}f^{(k)}),
\end{equation*} 
where $\mathcal{A}$ represents the Radon transformation, and $\mathcal{A}^T$ represents the backprojection. The iteration basically takes the projection image of the $k$th reconstruction outcome, evaluates its deviation from the input projection data, and finally uses the deviation to improve the reconstruction outcome \cite{Kak2001}. Observed from our experiments in Section \ref{sec:numerical1} as well as other literature \cite{Zeng2012_2}, the SIRT iteration tends to overfit the observation noises contained in the input sinogram as $k$ increases. Therefore, the determination of an appropriate iteration number is important, but there is no good rule of thumb for determining it. To mitigate the overfit issue, Wolf et al. \cite{Wolf2014} proposed to replace the backprojection operator $\mathcal{A}^T$ with a filtered backprojection using a low-pass filter, which we refer to as the filtered SIRT (fSIRT) in this paper. Motivated by the same idea, we propose to replace the backprojection operator $\mathcal{A}^T$ with our sFBP which has greater denoising capabilities than the existing FBP methods as shown in Section \ref{sec:numerical1}. We also use a stopping criteria of the iteration to avoid unnecessarily long iterations and consequently reduce the computation time for reconstruction. We stop the iteration if the following convergence criteria is met,
\begin{equation} \label{eq:stop}
|f^{(k+1)} - f^{(k)}| \le \epsilon.
\end{equation}
The SIRT iteration with the new filtered backprojection is referred to as `Sparse Filtered SIRT (sfSIRT)'. We conjecture that the proposed method converges faster than the standard SIRT, because sFBP regularizes the backprojection operation, so the resulting reconstruction outcome is less fluctuating. The faster convergence implies a less number of the iterations required until convergence so a less computation time. We will use numerical examples to show this in the next subsection.

\subsection{Reconstruction Accuracy for Limited Angle Scenarios} 
\begin{figure}
\centering
\includegraphics[width=1.1\linewidth]{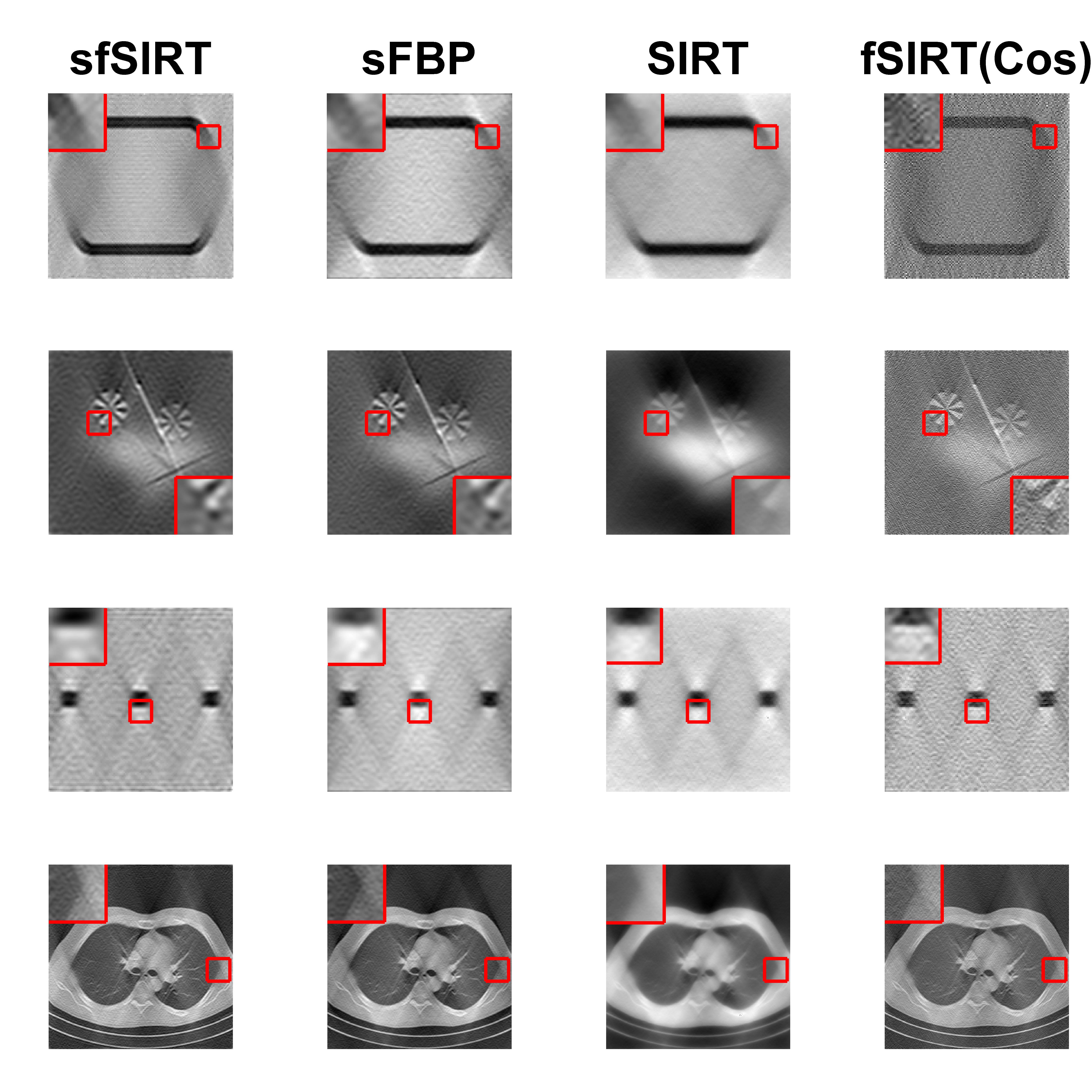}
\caption{Reconstruction of the sfSIRT, sFBP, SIRT and fSIRT with Cosine filter for $P=10^{5.5}$ and $r=65$. Each panel contains a sub-figure that illustrates the magnified image of a portion of the reconstruction result; the portion magnified was marked with a red box.}\label{fig:test2}
\end{figure}

For a numerical study, we emulated multiple limited angle scenarios with different choices of tilt angle ranges $(-r, r)$, while fixing the Poisson noise intensity at $P=10^{5.5}$. We varied $r \in \{65, 70, 75, 80, 85, 90\}$. For each choice of $r$, sinogram data was generated for each integer-valued tilt angle belonging to the interval $(-r, r)$, following the same data generation procedure in Section \ref{sec:numerical1}. The sinogram data were used as an input to the compared methods, which include sFBP, FBP, SIRT, fSIRT, L-FBP, TR, and sfSIRT (our method). For FBP, L-FBP and fSIRT, we applied the Cosine filter instead of the Hann filter and the Ram-Lak filter, because the Cosine filter produced better results in our numerical trials. For SIRT, sfSIRT, fSIRT and TR, the maximum number of iterations performed is 100; the iteration may stop earlier because we applied the stopping criterion \eqref{eq:stop}. The PSNR and SSIM measures of the methods were measured for 10 replicated simulation cases, and the measures were averaged over the replications. \\
\\
The PSNR and SSIM performances of the methods are summarized in Figure \ref{fig:test2}. In terms of PSNR, our proposed method (sfSIRT) outperformed the other competing methods for most of the experimental settings, and the proposed method was consistently the top three performers in terms of SSIM. In particular, sfSIRT performed better than sFBP significantly (a non-iterative version of our proposed method), which was performed worse than SIRT for limited angle scenarios. This says that iteratively applying sFBP within SIRT is quite effective in mitigating the missing wedge effect. In addition, sfSIRT uniformly outperformed SIRT in PSNR, which implies that introducing the proposed sFBP in SIRT iterations gives additional performance gains over SIRT. For qualitative comparison, we present the reconstruction outcomes of sfSIRT, fSIRT, SIRT and sFBP in Figure \ref{fig:test2}. In the figure, each sub-panel contains the reconstruction outcome of one of the compared methods with the magnified image of a portion of the reconstruction; the magnified portion was chosen to the area where the missing wedge effect is significant. The proposed method reduces the image artifacts due to the missing wedge. For example with the three dot phantom image, the white spots below the three dots were the image artifacts, and the sfSIRT reduced the artifacts significantly. The sfSIRT is still limited to the box phantom image, but this is the same for all compared methods.

\begin{figure}
\begin{center}
\includegraphics[width=1.2\textwidth]{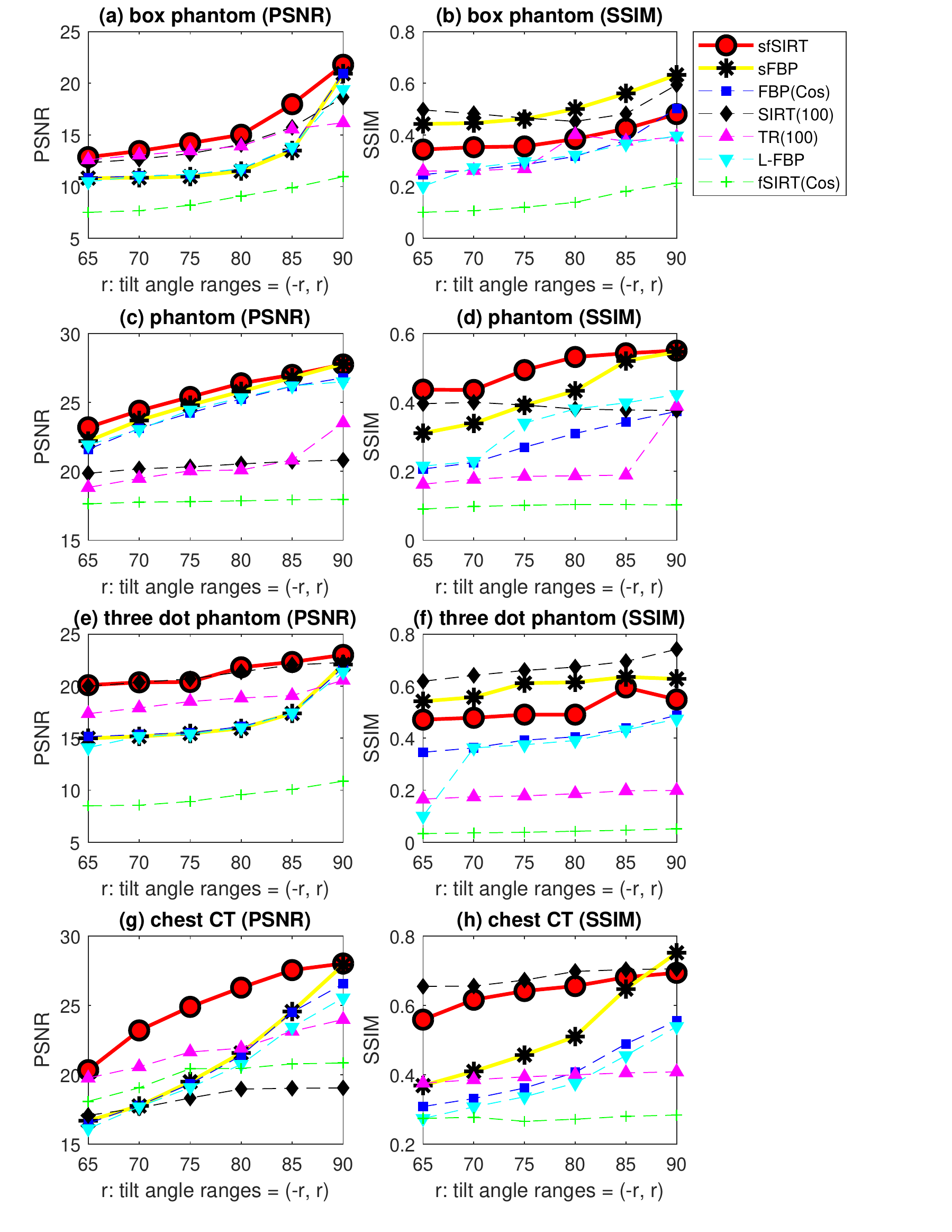} 
\caption{PSNR and SSIM of reconstructions for different ranges of available tilt angles.}\label{fig:test2_psnr}
\end{center}
\end{figure}

\subsection{Comparison of Computation Time}\label{sec:cca}
In this section, we compare the computational costs of three iterative tomography reconstruction methods, SIRT, fSIRT and our proposed sfSIRT. Since their computation costs per iteration are comparable, the computation costs of the three methods largely depend on the number of the iterations run, which are determined by how fast the iterative methods converge. We compare the computation times and convergence behaviors of the three methods.\\

For the comparison, we use the fourth test image, thorax CT image, as benchmark data, and ten noisy sinograms of the benchmark image were simulated using the same way that we used in the previous section with the settings $P=10^{5.5}$ and $r=65$. We applied the stopping criterion \eqref{eq:stop} for all of the three iterative methods. Table \ref{tab:time} summarizes the average number of iterations run and the corresponding computation times on average. The sfSIRT has converged faster that SIRT; the sfSIRT converged in 8.4 iterations on average, while SIRT converged in 20 iterations. The reason for the faster convergence of the sfSIRT can be explained by its use of sFBP as the backprojection operator in the SIRT iteration. The sFBP optimizes its backprojection filter with some regularization, and accordingly the iteration outcome of sfSIRT is regularized. This regularization on the solution path over iterations leads to the overall faster convergence for sfSIRT, while there is no such regularization on SIRT. \\

\begin{table}[t]
\renewcommand{\arraystretch}{1.1}
\caption{Average computation times (seconds) and the average number of iterations run under $P=10^{5.5}$ and $r=65$ for the fourth test image.}
\label{tab:time}
\centering
\begin{tabular}{c|c|c}
\hline
Methods & Time (seconds) & \# of iterations to stop \\ \hline
SIRT with stopping condition & 26.1231 & 20.0 \\
fSIRT with stopping condition & Do not converge & N.A. \\
sfSIRT with stopping condition & 11.54 & 8.4\\
sFBP & 2.12808  & N.A. \\
L-FBP & 2.20 & N.A.\\
SMF-FBP & 5.73 & N.A.\\\hline
\end{tabular}
\end{table}

\begin{figure*}
\begin{center}
\includegraphics[width=1.1\textwidth]{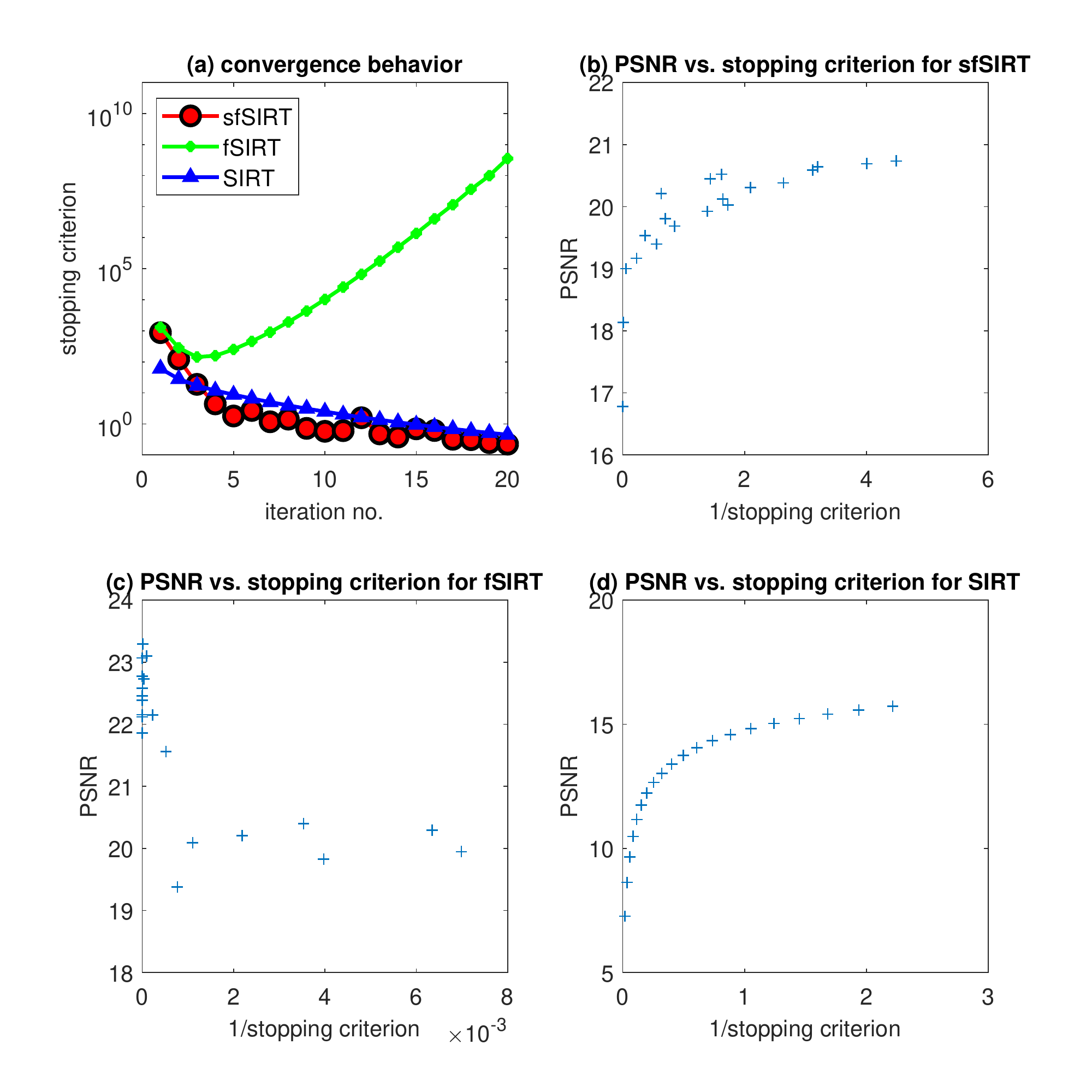} 
\caption{Convergence Behavior of sfSIRT (our method), fSIRT (with Cosine filter) and SIRT.}\label{fig:conv}
\end{center}
\end{figure*}

To look at the convergence behavior of the three method, we plotted the stopping criterion values over iterations in \ref{fig:conv}-(a). The stopping criterion for sfSIRT decreases faster than those for the other two methods, and the value actually diverged for fSIRT. We also looked at how the stopping criterion is related to PSNR values in Figure \ref{fig:conv}-(b), (c) and (d). For both of sfSIRT and SIRT, PSNR values were inversely correlated to the stopping criterion. This implies that stopping at lower stopping criterion would give better PSNR performance, so the choice of the stopping criterion gives a reasonable guidance to acquire a better reconstruction. However, for fSIRT, the correlation was random, so the stopping criterion does not really give a good guide to the reconstruction accuracy, which may limit the applicability of fSIRT. 

\section{Real Data Examples}\label{sec:3d}
\begin{figure}[p]
\begin{center}
\includegraphics[width=1.5in]{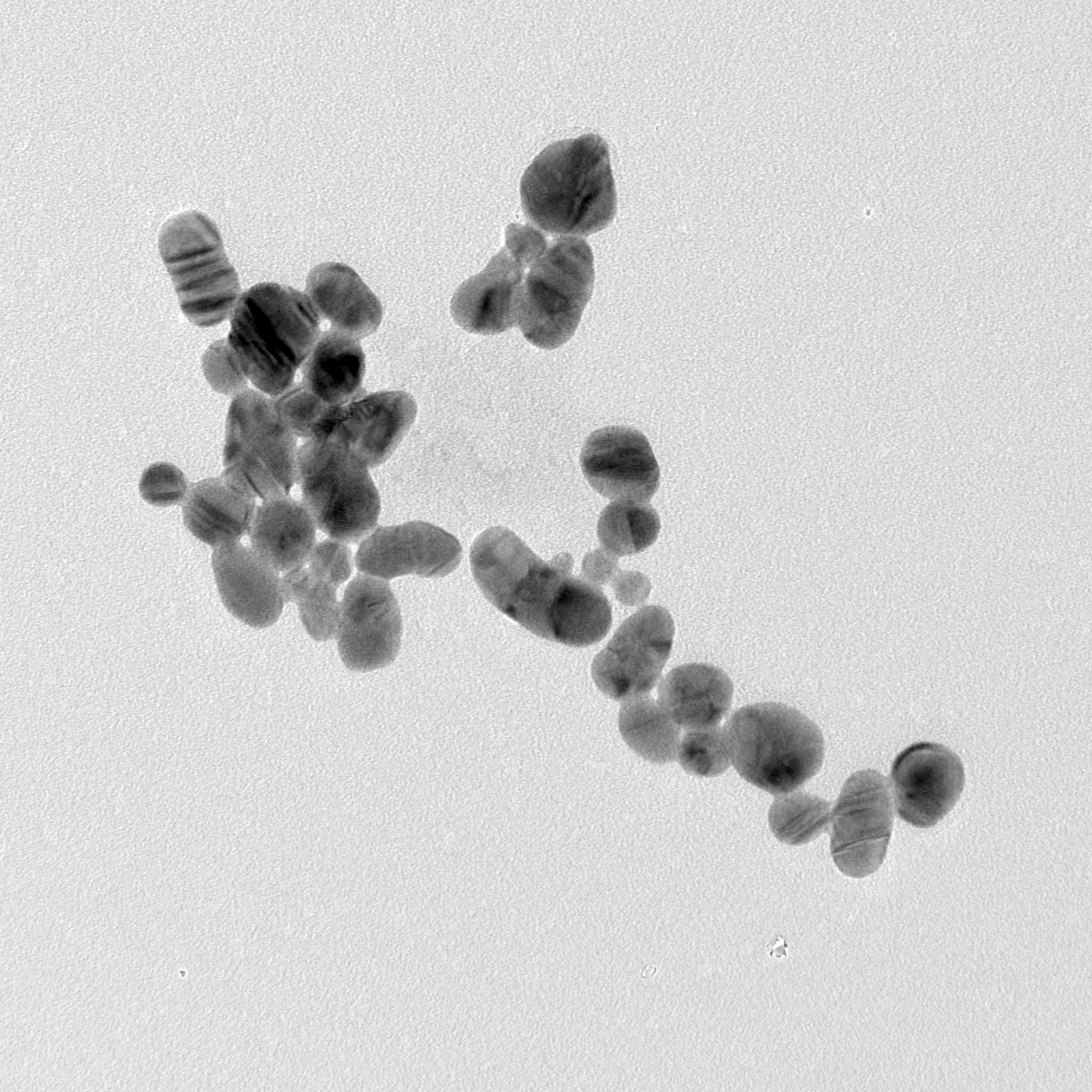} 
\caption{64th projection image of the gold nanoparticle tomography data} \label{fig:np64}
\end{center}
\begin{center}
\includegraphics[width=\linewidth]{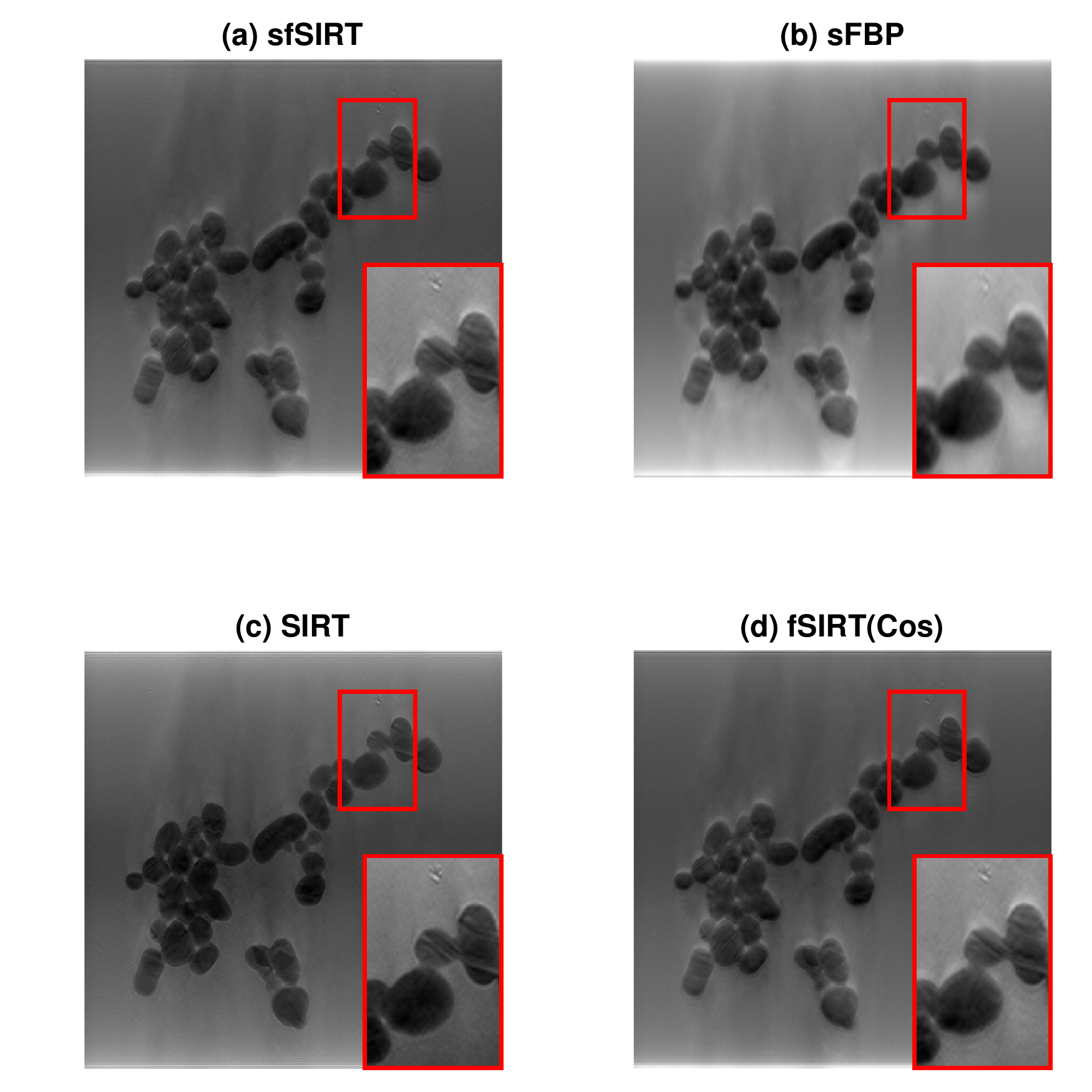} 
\caption{3D reconstructions results from the gold NP tomographic reconstructions from (a) sfSIRT (b) sFBP (c) SIRT (d) fSIRT with Cosine filter. The reconstruction viewed through x-y plan was shown, and each sub-figure comes with a partial magnification of the sub-figure on its bottom right. } \label{fig:3dproj}
\end{center}
\end{figure}

We apply sfSIRT, sFBP, SIRT and fSIRT with Cosine filter to electron tomography images we took for a sample of gold nanoparticles. The images consist of the projections from 103 different tilt angles in between $-65^{\circ}$ to $65^{\circ}$, in which from $-65^{\circ}$ to $0^{\circ}$ we get the one projection image roughly per $2^{\circ}$ and from $0^{\circ}$ to $65^{\circ}$ we get the one projection image roughly per $1^{\circ}$. Tomographic tilt-series were acquired manually using SerialEM \cite{Mastronarde2005} on a JEOL-ARM200cF cold field emission microscope operated at 200kV. A Fischione tomography holder was used, which can tilt up to 90 degrees. The images were obtained using a Gatan Orius 2k by 2k camera. As an example, the 64th projection image is shown in Figure \ref{fig:np64}. Each image was taken with a pixel size 0.13 nm by 0.13 nm and an acquisition time of 1 second. We used Etomo from IMOD \cite{Kremer1996} in doing the alignment of the raw TEM projection images. For tomography reconstruction, we used the `reconstruction-and-stacking' method as we described in Section \ref{sec:sfbp}, which reconstructs the 2D slices of a tomogram independently using the corresponding slices of sinograms. \\

The results of the 2D reconstructions are shown in Figure \ref{fig:3dproj}. Each sub-figure presents the 2D reconstruction viewed through the $(x,y)-$plane, which comes with the magnified image of a part of the reconstruction. For this dataset, the image artifact due to the missing wedge is the white spots widely spread around actual gold nanoparticles (shown as dark-colored) along the y-direction. The sfSIRT reduces the white spots significantly.

\section{Conclusion}\label{sec:con}
In this paper, we present a novel filter-optimization approach to improve the filtered backprojection (FBP) for tomographic reconstruction. The new approach, sFBP, formulates an optimization problem that optimizes the backprojection filter to minimize a regularized reconstruction error. The optimal solution of the optimization formulation can be achieved by applying a simple thresholding rule on the one-dimensional Fourier transform of sinogram data. This simple solution approach makes the computation of sFBP is as fast as that of the conventional FBP. In many numerical examples with simulated noisy sinogram, the approach has shown the greater trade-offs between the reconstruction accuracy and the computation efficiency than other existing methods. In particular, its reconstruction accuracy for higher noise cases was superior to the compared methods. \\

We also proposed the use of sFBP as a plug-in backprojection operator within the simultaneous iterative reconstruction technique (SIRT), which significantly improved the reconstruction accuracy over sFBP and SIRT when sinogram data are only available for a limited range of projection angles (or holder tilt angles in electron tomography). The computation time for the SIRT combined with sFBP, which we refer to as `sfSIRT', is in between those of sFBP and SIRT. It spends more computation time than sFBP, because sfSIRT needs to repeat sFBP multiple times to iteratively update its reconstruction. On the other hand, sfSIRT spends much less computation time than the conventional SIRT, because the sfSIRT's iteration converges faster than the SIRT iteration. We believe the proposed approach would be very practical with those strengths. The general idea has be easily generalized and modified to three-dimension tomographic reconstruction problems, which was shown using a real data example in 3D electron tomography.

\section*{Acknowledgment}
The authors would like to acknowledge support for this project. This work is partially supported by AFOSR FA9550-18-1-0144. The TEM work was performed at the National High Magnetic Field Laboratory, which is supported by NSF DMR-1157490 and the State of Florida. 

\bibliographystyle{elsarticle-num}
\bibliography{ofbpbib}
\end{document}